\documentclass{article} 
\usepackage[final]{colm2026_conference}
\usepackage[utf8]{inputenc}
\usepackage[T1]{fontenc}
\usepackage{microtype}
\usepackage{hyperref}
\usepackage{url}
\usepackage{booktabs}
\usepackage{amsfonts}       
\usepackage{nicefrac}       
\usepackage{amsmath}
\usepackage{amssymb}
\usepackage{xcolor}         
\usepackage{fancyhdr}       
\usepackage{graphicx}       
\usepackage{tcolorbox}
\usepackage{comment}
\usepackage{placeins}
\usepackage{float}
\usepackage{adjustbox}
\usepackage{subcaption}
\usepackage{natbib}

\usepackage{lineno}

\definecolor{darkblue}{rgb}{0, 0, 0.5}
\hypersetup{colorlinks=true, citecolor=darkblue, linkcolor=darkblue, urlcolor=darkblue}

\title{In-context superposition: human-like working memory interference in large language models}

\author{Hua-Dong Xiong$^{1}$, 
Li Ji-An$^{2}$, 
Jiaqi Huang$^{3,4}$, 
Robert C. Wilson$^{1,5}$,  
\AND Kwonjoon Lee$^{4}$\thanks{Co-senior author}, 
Xue-Xin Wei$^{6}$\footnotemark[1]
\\[1.5em]
$^{1}$ School of Psychological and Brain Sciences, Georgia Tech \\
$^{2}$ Department of Psychology, New York University \\
$^{3}$ Department of Cognitive Science, Indiana University Bloomington \\
$^{4}$ Honda Research Institute USA \\
$^{5}$ Center of Excellence for Computational Cognition, Georgia Tech\\
$^{6}$ Departments of Neuroscience and Psychology, The University of Texas at Austin
}

\begin{document}

\ifcolmsubmission
\linenumbers
\fi

\maketitle

\begin{abstract}

Intelligent systems must maintain and manipulate task-relevant information online to adapt to dynamic environments. This capacity, known as working memory, is fundamental to human reasoning. Yet, human working memory is strikingly limited, maintaining only three to four items in a brain with billions of neurons. Surprisingly, large language models (LLMs), despite different substrates and direct access to prior context through attention, exhibit similar working memory limitations. Why should such different systems face analogous constraints? We propose that working memory limitations reflect a general trade-off of shared representations: representational compression and reuse support efficient learning and generalization, but also cause simultaneously active representations to interfere. We show a two-layer transformer trained on a working memory task can solve it perfectly, but diverse trained LLMs exhibit human-like limitations: performance declines with memory load, while retrieval is biased by recency and stimulus statistics. Mirroring humans, working memory performance in LLMs is also associated with broader model capability. Mechanistically, we show that LLMs encode multiple memories in entangled representations --- a condition we call \emph{in-context superposition} --- and progressively suppress competing content while aligning the target with the readout. Moreover, a causal intervention that suppresses interfering information improves performance. Together, these findings suggest that working memory capacity reflects the ability to select task-relevant information under interference, a computational challenge shared by biological and artificial systems.

\end{abstract}

\section{Introduction}
\label{sec:intro}

Storing and processing task-relevant information is essential for adapting to changing environments and meeting diverse task demands. This function, known as working memory in psychology and neuroscience, is nevertheless strikingly limited: humans can maintain only a small number of items at once \citep{miller_magical_1956, cowan_magical_2001, ma_changing_2014}. This limitation is particularly puzzling because working memory is central to cognition, with individual differences in working memory capacity strongly associated with reasoning \citep{kyllonen_reasoning_1990} and general intelligence \citep{carpenter_what_1990, ackerman_working_2005, conway_working_2003}. This challenge is not unique to biological brains: although LLMs differ radically from biological brains in substrate, they nevertheless exhibit similar working memory limitations \citep{gong_working_2024, langis_strong_2025}. Given both its functional importance and the vast computational resources available—billions of neurons in the human brain and direct access to prior context through self-attention in LLMs—why should working memory remain so severely constrained and become a bottleneck for broader cognitive ability?

A prominent view in psychology and neuroscience is that working memory is limited not purely by representation capacity, but by interference control: intermediate representations must be maintained while irrelevant or competing alternatives are prevented from controlling subsequent computation \citep{cohen_control_1990, botvinick_conflict_2001, engle_working_1999, kane_workingmemory_2003, buschman_balancing_2021}. The same computational demand recurs in complex reasoning. Working memory capacity may therefore reflect a broader ability to resolve representational competition. But why should interference be so pervasive in intelligent systems with vast representational capacity?

The set of possible states, features, relations, and contexts in the world that an intelligent system may need to represent is combinatorial, making it impractical to assign an independent representational dimension to every variable. Instead, intelligent systems must compress representations and reuse representational structure across shared dimensions. Prior work has shown that neural networks can encode multiple features in superposition within a shared representational space \citep{elhage_toy_2022}. Such representational sharing can support efficient learning and generalization \citep{caruana_multitask_1997, musslick_multitasking_2017}, but it also carries an interference cost. When multiple variables are simultaneously active, shared or non-orthogonal representations can allow task-irrelevant variables to influence ongoing computation. Working memory limits may therefore reflect a trade-off inherent to shared representations: representational reuse supports efficient learning and generalization but can create interference when several variables are active and only one is relevant to the current computation.

Prior work on superposition has primarily examined how multiple learned features share a representational space \citep{elhage_toy_2022}. Working memory poses an analogous problem online: multiple items coexist in active, entangled representations, such that successful recall depends on selecting task-relevant information while suppressing competing content. We call this condition \emph{in-context superposition}. Under this view, humans and LLMs, despite their radically different substrates, both face a similar computational problem: maintaining multiple contextual variables in active representations while controlling which of them guide ongoing computation.

We tested this account in the N-back task, a canonical working memory paradigm, using behavioral manipulations, attention profiles, representational analyses, and a targeted intervention. Although a two-layer transformer trained specifically on N-back can solve the task perfectly, a diverse set of trained LLMs shows systematic performance declines with increasing memory load. First, retrieval errors vary systematically with recency, lure structure, stimulus set size, and transition statistics, showing that non-target variables influence the response and reproducing interference signatures observed in human working memory. Second, target-selective attention heads retain access to competing recent items, while multiple memory items remain encoded in entangled internal representations, revealing in-context superposition. Third, across layers, LLMs progressively suppress competing content and align the task-relevant target with the readout; variation in these neural measures predicts recall, and an optimized intervention sweep identified configurations in which reducing letter-identity information improved retrieval. Finally, working memory performance across LLMs is associated with broader model capability, paralleling the relationship between working memory capacity and broader cognitive ability in humans. Together, these findings link working memory performance to the control of interference among active contextual representations.

\section{LLMs show human-like working memory interference signatures}
\label{sec:behavioral}

\paragraph{Task.} We adapted the human N-back task for LLMs as a multi-turn interaction. Each trial comprised 50 turns. At turn $t$, the user supplied a single letter, and the model was required to produce the letter presented $N$ turns earlier (Fig.~\ref{fig:perf_task}a). For instance, in a 2-back trial with inputs \texttt{A,B,C,D,...}, the correct outputs are \texttt{-,-,A,B,...}, where each response replicates the input from two turns prior; the initial $N$ responses are \texttt{-}, reflecting insufficient context. Input letters were sampled uniformly from the 26-letter English alphabet (details in Appendix~\ref{app:prompts}). We evaluated 200 independent trials per condition for $N = 1,\ldots,4$.

\paragraph{Models.} We evaluated instruction-tuned models from four families, across a range of model sizes: Gemma~3 \citep{gemma_gemma_2025} (1B, 4B, 12B, 27B) and Qwen~3.5 \citep{qwen_qwen35_2026} (2B, 4B, 9B, 27B), allowing within-family scaling analysis. For cross-family comparison, we included Llama-3.1-8B-Instruct \citep{grattafiori_llama_2024} and Ministral~3 14B \citep{liu_ministral_2026}. In total, we analyzed 10 models; the complete model list is reported in Appendix~\ref{app:models}.

\paragraph{Evaluation procedure.} We report results under both teacher-forced and autoregressive conditions. For autoregressive mode, we used each model family’s default generation settings. In teacher-forced mode, the assistant was conditioned on the ground-truth answers from previous turns rather than on its own prior outputs. All models were evaluated in non-thinking mode.

\subsection{LLMs show load-dependent limits in working memory capacity}

\begin{figure}[htbp]
\centering
\includegraphics[width=0.9\textwidth]{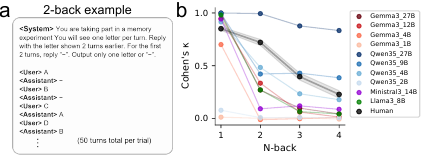}
\vspace{-5pt}
\caption{\textbf{LLMs' N-back performance decreases with increasing memory load.} \textbf{(a)} Each trial proceeds as a multi-turn dialogue: the user provides one letter per turn, and the model outputs the memory item presented $N$ turns earlier, emitting \texttt{-} when insufficient context is available. \textbf{(b)} Under autoregressive mode, Cohen's $\kappa$ declines with increasing memory load $N$. This metric normalizes performance across humans and models, with 0 indicating chance level and 1 indicating perfect performance. The shaded region for humans denotes the 95\% bootstrap confidence interval of the aggregated human reference curve.}
\label{fig:perf_task}
\end{figure}

In principle, the N-back task could be solved by a sufficiently clean positional retrieval strategy. Because transformers can access the full token history, and rotary positional embeddings \citep{su_roformer_2024} make relative offsets especially accessible, a model could identify the memory item at the required position and read out its content with minimal interference. Under such a solution, one would not expect a working memory capacity limit. To verify that the architecture is sufficient, we trained a two-layer, single-head causal transformer with rotary positional embeddings directly on the N-back task. This model achieved perfect accuracy on a held-out test set, establishing the architectural sufficiency of even a shallow transformer for N-back (details in Appendix~\ref{app:toy_transformer}). We then asked whether trained LLMs, which have the same architectural access to prior context, nevertheless exhibit working memory limitations, as previously reported \citep{gong_working_2024, langis_strong_2025}.

In contrast, all trained LLMs we tested exhibited limited working memory capacity (Fig.~\ref{fig:perf_task}b, Fig.~\ref{fig:retrieval_ar}a). We report Cohen’s $\kappa$ rather than raw accuracy because the human and LLM task variants have different chance levels; $\kappa$ therefore quantifies performance above random guessing. Raw accuracies by model and evaluation mode are reported in Appendix Table~\ref{tab:nback_accuracy}. Human performance was calculated by pooling data from multiple N-back studies (details in Appendix~\ref{app:human_aggregation}). As memory load $N$ increased, performance dropped (Fig.~\ref{fig:perf_task}b): most models approached chance by 3- or 4-back and often fell below human levels. Although Qwen~3.5 27B exceeded human performance, its accuracy still declined with $N$, preserving a qualitatively human-like capacity limit that is difficult to reconcile with a clean positional retrieval account --- especially given the very small offsets involved. This qualitative pattern also holds under teacher forcing (Appendix Fig.~\ref{fig:capacity_tf}), showing that increasing load continues to impair the selection of task-relevant information even when prior responses are correct and the relevant memory remains available.

\subsection{Recency interference in LLMs' working memory}

\begin{figure}[htbp]
\centering
\includegraphics[width=\textwidth]{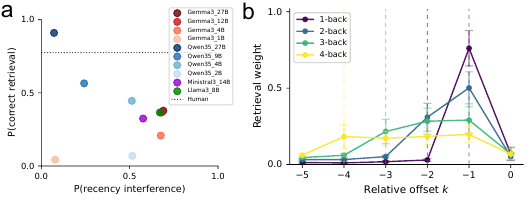}
\vspace{-5pt}
\caption{\textbf{Recency interference in LLM working memory behaviors.} \textbf{(a)} Across models, higher target retrieval probability is associated with lower recency interference, revealing a capacity--recency relationship. \textbf{(b)} Temporal retrieval kernels assign substantial probability mass to recent non-target memory items. Error bars indicate $\pm$1 standard error of the mean (SEM) across models.}
\label{fig:retrieval_ar}
\end{figure}

Recency error patterns can distinguish a position-dominant retrieval account from an interference-based explanation. In both accounts, positional information may help identify the target. The key difference is whether retrieval is driven primarily by isolating the correct relative position, or whether non-target memory content remains sufficiently active to compete with the target at readout. If the latter holds, retrieval errors should be systematically biased toward recent non-target memory items, consistent with classic interference effects in human working memory \citep{baddeley_recency_1993, greene_sources_1986, howard_distributed_2002}.

We index relative position by a signed offset $k \leq 0$, with $k=0$ denoting the current stimulus and $k=-N$ the target. We define recency interference as probability mass assigned to non-target stimuli at offsets $k=-N+1,\ldots,0$ --- memory items that remain active but are not the correct response on the current turn. Across models, higher target retrieval probability was associated with weaker recency interference, revealing a capacity--recency relationship (Fig.~\ref{fig:retrieval_ar}a). The temporal retrieval-probability kernel --- the probability mass assigned across recent relative offsets --- further showed disproportionate mass on recent non-target items, and this recency bias grew with memory load (Fig.~\ref{fig:retrieval_ar}b). Teacher forcing mitigated part of this effect, but the same qualitative pattern persisted (Appendix Fig.~\ref{fig:retrieval_tf}). Together, these retrieval profiles show that increasing load changes not only whether retrieval succeeds, but how competing memories systematically shape the resulting choice, consistent with interference from concurrently active memory items rather than a purely positional lookup.

\subsection{Content-based interference biases LLMs' retrieval}

If retrieval is influenced by multiple recent memory items rather than by position alone, it should also be sensitive to stimulus content, producing systematic effects of lure similarity, set size, and transition structure.

Human N-back studies show several such content-based signatures, including lure effects and stimulus set-size effects \citep{conway_working_2003, engle_working_1999}. We tested three manipulations in LLMs. First, we introduced $N \pm 1$ lure stimuli: with probability $p_{\mathrm{lure}} = 0.25$, the current input was replaced by the memory item at $t\!-\!(N\!-\!1)$ or $t\!-\!(N\!+\!1)$. Lure stimuli test whether content-matching items already in memory --- one already recalled ($t\!-\!(N\!+\!1)$) or one to be recalled next ($t\!-\!(N\!-\!1)$) --- interfere with retrieval of the current target. Second, we reduced the active set size from 26 letters to 10 per trial, increasing the probability of content repetition. Third, we imposed sequential transition structure by sampling 10 letters and arranging them in a circular Markov chain, such that each letter was followed by its successor with probability $p_{\mathrm{tran}} = 0.8$. Under a position-only strategy, these manipulations should have limited effect; under content-sensitive retrieval, they should systematically bias performance.

We hypothesize that models showing larger interference effects under these manipulations would have weaker interference control and therefore lower working memory capacity on the N-back task. We quantified each interference effect as the signed change in performance produced by a given manipulation relative to the baseline condition. We then correlated each signed condition effect with N-back accuracy and interpreted its direction separately for each manipulation (Table~\ref{tab:interference_corr}). As predicted, larger performance drops under lure presentations and reduced stimulus set sizes correlated significantly with lower N-back accuracy across models (Table~\ref{tab:interference_corr}). Interestingly, the effect of transition statistics differed by evaluation mode. Under autoregressive generation, being more strongly influenced by sequence regularities improved performance, because a content-based prediction shortcut often aligned with the correct answer in the structured condition and therefore succeeded 80\% of the time. Under teacher forcing, where the correct retrieval path was already available, being more strongly influenced by the same sequence regularities predicted worse performance, indicating less effective suppression of task-irrelevant statistical structure.

Across recency, lure, vocabulary, and transition manipulations, retrieval errors are structured by non-target content that remains behaviorally influential despite being task-irrelevant under a clean positional rule. Together, these signatures parallel classic interference phenomena in human working memory and demonstrate that LLM working memory limits reflect competition among simultaneously represented contextual items.

\begin{table}[t]
\centering
\caption{\textbf{Interference manipulations predict N-back accuracy across models.} Pearson $r$ between each signed condition effect and accuracy, pooled across models and $N$ levels. $^{*}p<.05$; $^{***}p<.001$.}
\label{tab:interference_corr}
\small
\begin{tabular}{l r@{.}l r@{.}l r@{.}l}
\toprule
& \multicolumn{2}{c}{\textbf{Lure}} & \multicolumn{2}{c}{\textbf{Stimulus size}} & \multicolumn{2}{c}{\textbf{Transition statistics}} \\
\midrule
Autoregressive  & $-$&51$^{***}$ & $-$&30$^{*}$ & &36$^{*}$ \\
Teacher forcing & $-$&32$^{*}$   & $-$&26   & $-$&62$^{***}$ \\
\bottomrule
\end{tabular}
\end{table}

\subsection{Target-selective heads retain attention to competing memory items in representative models}

\begin{figure}[htbp]
\centering
\includegraphics[width=\textwidth]{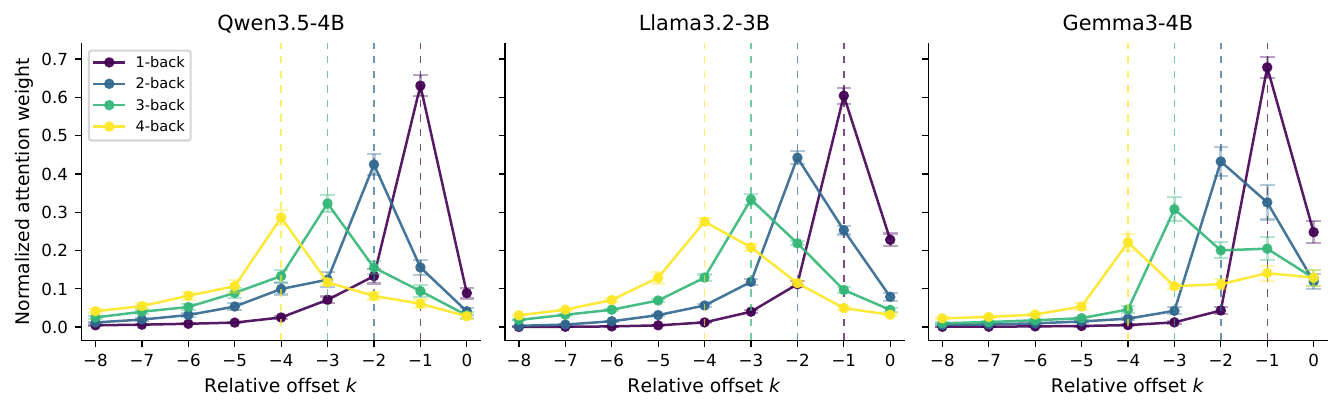}
\vspace{-5pt}
\caption{\textbf{Target-selective attention heads retain broad recency profiles.} For each model and $N$-back level, we identified the 10 heads with the strongest preference for the target position in teacher-forced baseline trials and plotted their mean normalized attention to previous stimulus tokens as a function of relative offset $k$. Dashed lines mark the target offset $k=-N$. Error bars indicate $\pm$1 SEM across heads.}
\label{fig:attention_lag_profiles}
\end{figure}

Target-selective attention routed both the target and competing recent memory items. We analyzed three representative models for which explicit pairwise attention weights could be extracted: Qwen3.5-4B, Llama3.2-3B, and Gemma3-4B. In teacher-forced baseline trials, we measured the attention row at each response-prediction position, normalized attention to each previous stimulus token by the total attention assigned to observed stimulus tokens, and ranked heads by their selectivity for the target offset $k=-N$ (details in Appendix~\ref{app:attention_methods}). Strong target attention was concentrated in a small subset of heads (Appendix Fig.~\ref{fig:attention_head_survival}), primarily in middle-to-late layers (Appendix Fig.~\ref{fig:attention_head_depth}), and many of the same top heads recurred across $N$-back levels, especially in Qwen3.5-4B and Llama3.2-3B (Appendix Fig.~\ref{fig:attention_head_overlap}).

Even the most target-selective heads retained broad recency profiles. Their attention peaked at the correct offset $k=-N$, confirming that they track the task-relevant offset, while substantial normalized attention remained on neighboring and more recent stimulus tokens (Fig.~\ref{fig:attention_lag_profiles}). The target peak weakened from 1-back to higher loads as attention spread across recent non-target tokens. Thus, target-selective attention provides positional access while routing competing contextual items into the response computation.

\section{LLMs show a common interference-control trajectory under in-context superposition}
\label{sec:mechanistic}

The behavioral and attention results show that competing recent items continue to influence the response even when attention is selective for the target position. We therefore examined how LLMs transform these item representations across layers. Studies of sequential working memory in biological systems have identified a dynamic representational trajectory: multiple items are maintained in representations that become more orthogonal over time, while the item selected for retrieval becomes aligned with the downstream readout \citep{murray_stable_2017, stokes_dynamic_2013, xie_geometry_2022}. We tested whether LLMs follow an analogous trajectory during \emph{in-context superposition}, with recent items initially occupying overlapping representations and becoming more differentiated before the target aligns with the readout.

To study interference control mechanistically, we analyze the response representation $h_t^\ell$, extracted after the assistant token and before the answer token is generated on turn $t$ at layer $\ell$. Using the signed relative offset $k$ defined above, $x_{t+k}$ denotes the item at offset $k$, with $k=0$ denoting the current item and $k=-N$ the target. For letter $c$, we define $h_{k,c}^\ell = \mathbb{E}[h_t^\ell \mid x_{t+k}=c]$. The collection $\{h_{k,c}^\ell\}_c$ characterizes the representation at relative offset $k$. To quantify residual letter-identity information, we compare the current-item representations $h_{0,c}^\ell$ with the letter-identity representations $s_c^\ell$ constructed from isolated encoding and generation contexts. To quantify task-relevant structure, we compare representations grouped by stimulus letter across target and non-target offsets (details in Appendix~\ref{app:analysis}).

\subsection{Layerwise signatures of interference control in LLMs}

We analyze four metrics that characterize how task-relevant and task-irrelevant representations evolve across layers. Letter alignment and inter-class decodability assess how much task-irrelevant stimulus-content information remains in the response representation. Subspace similarity measures the degree of overlap among active representations across different relative positions. Target alignment quantifies whether task-relevant information has been transformed into a form that effectively guides output readout.

\begin{figure}[htbp]
\centering
\includegraphics[width=\textwidth,clip]{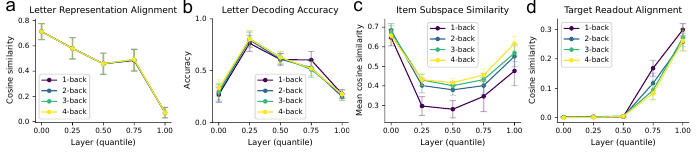}
\vspace{-5pt}
\caption{\textbf{A common layerwise interference-control trajectory across LLMs.} \textbf{(a)}~Alignment between the current-item representations $h_{0,c}^\ell$ and the letter-identity representations $s_c^\ell$ declines across layers, indicating progressive suppression of letter-identity information. \textbf{(b)}~Decodability across current-item representations first increases in mid layers and then decreases toward the readout stage, consistent with gradual suppression of task-irrelevant stimulus identity prior to readout. \textbf{(c)}~Relative-position representations first become more separated through middle layers as subspace similarity declines, then partially overlap again in late layers as the network prepares the target for readout; higher $N$ retains greater overlap throughout, showing that temporal positions remain only partially disentangled. \textbf{(d)}~Alignment between the centered target representations and the correct readout directions remains near zero until late layers, then rises sharply as the representations become more directly readable by the output layer. Error bars indicate $\pm$1 SEM across models.}
\label{fig:neural_control}
\end{figure}

Both letter alignment and inter-class decodability persist into the network. Alignment between current-item representations $h_{0,c}^\ell$ and letter-identity representations $s_c^\ell$ declines over successive layers, while inter-class decodability first rises in early layers before declining (Fig.~\ref{fig:neural_control}a,b). Grouping response states by stimulus letter for each relative offset $k \in \{-N,\ldots,0\}$ reveals that memory representations are initially similar, become more separated through the middle of the network as subspace similarity declines, then partially re-converge in late layers (Fig.~\ref{fig:neural_control}c). This trajectory is consistent with reduced interference as memory representations become more distinct through middle layers. Late layers nevertheless retain substantial overlap among relative-position representations --- especially under higher memory loads --- leaving target and non-target items entangled in the same activation space, consistent with resource-based models of working memory \citep{ma_changing_2014}. Finally, the target-alignment analysis reveals that alignment between the centered target representations and their corresponding readout directions remains low throughout early and middle layers, then rises sharply near the end of processing (Fig.~\ref{fig:neural_control}d). Teacher forcing reproduced the same representational sequence (Appendix Fig.~\ref{fig:neural_control_tf}), demonstrating that the trajectory persists when prior responses are fixed to ground truth.

\subsection{LLM representations co-vary with recall}

Recall should co-vary with this layerwise trajectory: sequence positions with more successful recall should exhibit reduced residual content, weaker similarity among relative-position representations, and stronger target alignment in late processing.

Across all $N$-back levels and models, stronger target alignment was consistently associated with better performance (Table~\ref{tab:trial_corr}). In contrast, higher levels of residual content—reflected in greater letter alignment and higher inter-class decodability—were associated with poorer recall. Similarly, greater similarity among relative-position representations co-varied with poorer recall.

\begin{table}[t]
\centering
\caption{\textbf{Representational metrics of interference control co-vary with recall.} Per-model Pearson $r$ between each metric (cf.\ Fig.~\ref{fig:neural_control}a--d) and per-position recall (mean target log-probability), pooled over $N = 1$--$4$ and aggregated across the ten models by Fisher $z$-transformation. $^{*}p<.05$; $^{***}p<.001$.}
\label{tab:trial_corr}
\small
\begin{tabular}{l r@{.}l r@{.}l r@{.}l r@{.}l}
\toprule
& \multicolumn{2}{c}{\textbf{Letter alignment}} & \multicolumn{2}{c}{\textbf{Decodability}} & \multicolumn{2}{c}{\textbf{Subspace similarity}} & \multicolumn{2}{c}{\textbf{Target alignment}} \\
\midrule
Autoregressive  & $-$&15$^{***}$ & $-$&42$^{***}$ & $-$&16$^{***}$ & &78$^{***}$ \\
Teacher forcing & $-$&13$^{***}$ & $-$&14$^{***}$ & $-$&22$^{***}$ & &76$^{***}$ \\
\bottomrule
\end{tabular}
\end{table}

\subsection{Optimized suppression of letter-identity directions can improve N-back performance}

\begin{figure}[htbp]
\centering
\includegraphics[width=0.72\textwidth]{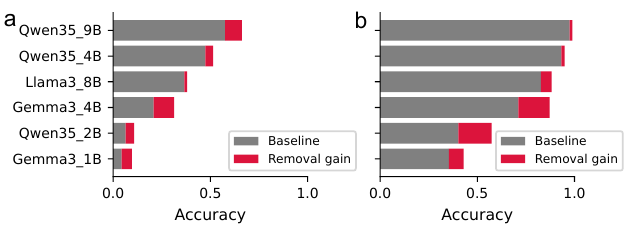}
\vspace{-5pt}
\caption{\textbf{Optimized suppression of letter-identity directions can improve N-back performance.} \textbf{(a)}~Autoregressive evaluation. \textbf{(b)}~Teacher-forced evaluation. For each model, gray bars show mean baseline accuracy across $N = 1$--$4$, and red segments show the mean accuracy gain from the best intervention found for each memory load in the causal removal sweep. The optimized gains are larger in weaker models, with limited headroom near ceiling.}
\label{fig:causal_removal}
\end{figure}

The analyses above suggest that residual letter-identity information is one source of interference during retrieval. We tested this with a direct intervention that selectively suppressed letter-identity directions in the residual stream. Specifically, we first averaged the letter-identity representations $s_c^\ell$ across the three contexts and then performed a singular value decomposition across letters to estimate letter-identity directions. We swept over individual principal directions, intervention strengths, and the two earliest sampled layer depths. The intervention acted only at the answer position of each turn --- the assistant-side hidden state immediately before the model generated the response token. We scored $N$-back performance on trials matched by random seed (details in Appendix~\ref{app:causal_remove}).

Across five LLMs, the best parameter configurations in our search produced aggregate N-back gains in both autoregressive and teacher-forced evaluation (Fig.~\ref{fig:causal_removal}). Gains were larger in weaker models, whereas stronger models had limited headroom near ceiling. The sweep therefore identified specific model and load conditions in which suppressing selected letter-identity directions improved retrieval. Because configurations were selected and evaluated on the same trials, generalization to held-out trials remains to be tested. The heterogeneous gains indicate that residual letter-identity information can contribute to interference alongside other mechanistic sources.

\section{LLMs' working memory performance is associated with broader model capability}

We found that, as in humans, working memory capacity in LLMs is associated with broader model capability. Mean N-back accuracy correlated significantly with MMLU Pro \citep{wang_mmlupro_2024}, GPQA Diamond \citep{rein_gpqa_2024}, and IFEval \citep{zhou_instructionfollowing_2023} under both autoregressive and teacher-forced evaluation (Table~\ref{tab:benchmark_corr}). This pattern mirrors a central finding from human cognition: individual differences in working memory capacity and interference control are linked to reasoning and intelligence ($r = 0.31$--$0.78, p < 0.05$) \citep{carpenter_what_1990, conway_working_2003, engle_working_1999}. As a baseline, log parameter count did not significantly predict any of the three benchmarks. The parallel association across two radically different systems is consistent with the view that interference control reflects a domain-general computational demand shared between working memory and complex reasoning.

\begin{table}[t]
\centering
\caption{\textbf{Working memory capacity is associated with broader model capability.} Pearson $r$ between each benchmark (MMLU Pro, GPQA Diamond, IFEval) and mean N-back accuracy (averaged over $N = 1$--$4$) across the 10 models evaluated, reported separately for autoregressive generation and teacher forcing. $^{*}p<.05$; $^{**}p<.01$; $^{***}p<.001$.}
\label{tab:benchmark_corr}
\small
\begin{tabular}{l r@{.}l r@{.}l r@{.}l}
\toprule
& \multicolumn{2}{c}{\textbf{MMLU Pro}} & \multicolumn{2}{c}{\textbf{GPQA Diamond}} & \multicolumn{2}{c}{\textbf{IFEval}} \\
\midrule
Autoregressive    & & 78$^{**}$  & & 88$^{***}$ & & 69$^{*}$ \\
Teacher forcing   & & 70$^{*}$   & & 72$^{*}$   & & 82$^{**}$ \\
\bottomrule
\end{tabular}
\end{table}

\section{Discussion}
\label{sec:discussion}

We investigated why trained LLMs exhibit limited working-memory capacity on N-back despite having full access to prior context and, in principle, sufficient architectural capacity to solve the task through direct positional retrieval. Across a diverse set of models, we found converging evidence against a position-only explanation. Behaviorally, performance declined with memory load, errors were biased toward recent non-target memory items, and retrieval was systematically modulated by lure similarity, stimulus set size, and transition statistics. Despite substantial variability in working-memory performance, LLMs converged on a common interference-control trajectory: task-irrelevant letter-identity information persisted in the residual stream and was progressively suppressed across layers; memory representations were reorganized into more distinct representational directions, consistent with reduced interference through middle layers; and target representations became aligned with the readout direction only late in processing. Together, these findings support the \emph{in-context superposition} hypothesis that representational interference is a substantial contributor to working-memory limitations in LLMs, while the optimized letter-identity suppression sweep identified conditions in which suppressing letter-identity information improved retrieval.

Retrieval combines positional access with interference control. The models have access to relative position, and the attention-analyzed models contain specialized heads that preferentially attend to the N-back target, but these heads also distribute attention across other recent items. The corresponding behavioral biases show that recent memories and content structure continue to shape retrieval. Successful retrieval therefore depends on both locating the target and suppressing task-irrelevant content to isolate it for readout.

This interpretation connects the present results to a broader computational perspective. Shared distributed representations support efficient learning and flexible generalization \citep{caruana_multitask_1997, musslick_multitasking_2017, elhage_toy_2022}, but when several contextual variables are active, they also create interference through the co-activation of overlapping features. Our results accord with this view: residual overlap, content-based retrieval biases, and progressive suppression of task-irrelevant content all indicate that working-memory capacity depends on interference control.

Humans and LLMs differ radically in substrate, architecture, learning rules, and developmental history, yet both face the same computational problem: retrieving task-relevant content from entangled representations while suppressing competing alternatives. This analogous demand may help explain why working-memory capacity correlates with general reasoning in both systems. Our findings further suggest that this challenge extends beyond N-back. Across models, better N-back performance was associated with stronger reasoning and instruction-following benchmarks, identifying interference control as a candidate broader dimension of model capability. Establishing the generality and causal role of this relationship will require a larger model set and direct intervention. Improving memory performance may therefore depend not only on expanding context access or retrieval bandwidth, but also on improving selective retrieval within shared representations --- a practical direction suggesting that interventions that better separate task-relevant from task-irrelevant content may benefit both short-horizon memory tasks and broader downstream capabilities.

\paragraph{Limitations and future work.}
Our use of \emph{in-context superposition} is primarily geometric: we establish partial overlap among relative-position-conditioned representations and relate this overlap to recall, while decomposing the response state into simultaneously represented item--position variables will determine whether this overlap reflects a dimensional bottleneck. Our experiments characterize inference-time behavior in non-thinking mode; allowing models to externalize intermediate computation may yield higher performance. The strongest scaling evidence comes from within-family comparisons across the model families studied; identifying the architectural or training determinants of interference control will require a larger and more diverse model set. N-back isolates a simple form of online selection under interference and is less structured than natural language reasoning, so observed variance may include instruction-following behavior or prompt-formatting effects. Finally, the intervention addressed letter-identity interference, while its heterogeneous gains point to additional representational and circuit-level constraints on selective retrieval.

Together, our results identify representational interference as a core constraint on working memory in trained LLMs: the critical challenge is not merely keeping information available in context, but selecting task-relevant information under interference.

\clearpage

\section*{Acknowledgments}
This work was supported by Lambda Research Grant awarded to H.-D.X. and Sloan Research Fellowship awarded to X.-X.W.

\section*{Reproducibility Statement}
Code for reproducing all experiments and analyses is available at \url{https://github.com/sakimarquis/in-context-superposition}.

\clearpage
\bibliography{references}
\bibliographystyle{colm2026_conference}

\clearpage
\appendix
\section{Appendix}

\subsection{Experimental Details}
\label{app:experimental}

\paragraph{Stimulus generation.}
Each trial contains 50 turns. At turn $t$, the user provides one stimulus letter $x_t$. In the baseline condition, each $x_t$ is drawn uniformly from the 26-letter English alphabet. In the reduced-vocabulary condition, 10 letters are first sampled for the trial, and each $x_t$ is then drawn uniformly from this subset. In the structured condition ($|\mathcal{V}| = 10$, $p_{\mathrm{tran}} = 0.8$), the same 10-letter subset is arranged in a circular loop (e.g., $A \!\to\! B \!\to\! C \!\to\! \cdots \!\to\! A$); at each turn, the next letter follows its loop successor with probability 0.8, and with probability 0.2 a letter is sampled uniformly from the remaining active letters (excluding the current letter and its loop successor). In the lure conditions ($p_{\mathrm{lure}} = 0.25$), the stimulus at turn $t$ is replaced with probability 0.25 by the memory item at $t\!-\!(N\!-\!1)$ or $t\!-\!(N\!+\!1)$, tested in separate runs. Across all conditions, each combination of condition and $N \in \{1, 2, 3, 4\}$ is evaluated over 200 independent trials.

\paragraph{Evaluation protocol.}
The task is presented as a multi-turn chat. A system prompt explains the $N$-back rule and provides worked examples (Appendix~\ref{app:prompts}). For a trial with stimulus sequence $(x_0, x_1, \ldots, x_{49})$, the correct response at turn $t$ is $y_t = \texttt{-}$ for $t < N$ and $y_t = x_{t-N}$ for $t \geq N$. In teacher forcing, when we score turn $t$, the context contains the user letters up to turn $t$ together with the correct assistant responses from all earlier turns. In autoregressive evaluation, those earlier assistant responses are replaced by the model's own outputs. We score accuracy only on turns $t \geq N$, counting a turn as correct when the model's top prediction matches $y_t$. For autoregressive analyses, we use each model family's default officially recommended generation settings rather than retuning decoding hyperparameters for this task.

\paragraph{Models.}
See Section~\ref{app:models} for a complete list of all models evaluated.

\paragraph{A two-layer transformer achieves perfect performance.}
\label{app:toy_transformer}
To verify that the task is solvable in principle with rotary position coding, we trained a small causal transformer with two transformer blocks, one attention head, model dimension 48, and dropout 0.1 on the baseline 50-turn N-back task. The model was optimized with cross-entropy on answer positions only, using AdamW (learning rate $3 \times 10^{-4}$, weight decay 0.01, batch size 128) for 100 epochs with 5 warmup epochs followed by cosine decay. Each epoch used 10{,}000 newly sampled training trials across $N \in \{1,2,3,4,6,8\}$, and evaluation used a fixed balanced test set of 200 trials per $N$, scored by top-1 accuracy on turns $t \geq N$. This model reached 100\% accuracy, confirming that a RoPE-equipped transformer can solve the task in principle.

\subsection{Human Reference Aggregation}
\label{app:human_aggregation}

The human reference band in Fig.~\ref{fig:perf_task} was constructed to pool heterogeneous letter-based N-back studies without collapsing them into a single participant-level dataset. We therefore aggregated at the study level: each study contributes one estimate per N-back level, and the final human reference at each level is the unweighted mean across studies. This choice preserves between-study variation in protocol and sample composition and avoids allowing any single large cohort to dominate the reference curve.

We combined five individual-level datasets with accuracy measurements in the 1--4-back range \citep{enkavi_largescale_2019, loffler_common_2024, burgoyne_which_2025, repovs_working_2012, shin_effects_2017}.

For individual-level studies, let $a_{i,s,n}$ denote participant $i$'s accuracy in study $s$ at N-back level $n$. For non-adaptive studies, we computed the study mean
\[
\bar{a}_{s,n} = \frac{1}{m_{s,n}} \sum_{i=1}^{m_{s,n}} a_{i,s,n},
\]
where $m_{s,n}$ is the number of participants observed in study $s$ at level $n$. Enkavi 2019 used an adaptive design, so higher N-back levels were observed only for participants who advanced from easier levels. To correct this selection bias, we fit a logistic progression model at each transition,
\[
\Pr(A_{i,j+1}=1 \mid d'_{i,j}) = \sigma(\beta_{0,j} + \beta_{1,j} d'_{i,j}),
\]
where $A_{i,j+1}$ indicates whether participant $i$ advanced from level $j$ to level $j+1$, and $d'_{i,j}$ denotes participant $i$'s sensitivity index $d'$ at level $j$. The probability that participant $i$ is observed at level $n$ is then
\[
\pi_{i,n} = \prod_{j=2}^{n-1} \Pr(A_{i,j+1}=1 \mid d'_{i,j}),
\qquad
w_{i,n} = \pi_{i,n}^{-1},
\]
and the IPW-corrected study mean for the adaptive study is
\[
\bar{a}^{\mathrm{IPW}}_{s,n} = \frac{\sum_i w_{i,n} a_{i,s,n}}{\sum_i w_{i,n}}
\]
for $n > 2$. For literature-only studies with no participant-level records, we used the reported study mean directly.

At each N-back level $n$, the final human reference was the equal-weight average across all available studies,
\[
\hat{\mu}_n = \frac{1}{|S_n|} \sum_{s \in S_n} \bar{a}_{s,n},
\]
where $S_n$ denotes the set of studies contributing data at level $n$, and $\bar{a}_{s,n}$ represents the study-level accuracy mean (using the IPW-corrected estimate $\bar{a}^{\mathrm{IPW}}_{s,n}$ for Enkavi 2019 when $n > 2$). Confidence intervals were estimated with 2{,}000 bootstrap resamples by resampling participants with replacement within each individual-level study, refitting the Enkavi progression model in each bootstrap sample, and sampling literature-only study means from a Gaussian distribution with the reported mean and standard error. For plotting in Fig.~\ref{fig:perf_task}b, the aggregated human accuracy estimates were then mapped to the chance-corrected scale used on the y-axis.

\subsection{Models}
\label{app:models}

All models were accessed via publicly released weights. Parameter counts are reported in billions (B). We evaluated all models in non-thinking mode. For the Qwen~3.5 family, which natively supports chain-of-thought reasoning, thinking mode was explicitly disabled so that all models were evaluated under comparable non-thinking conditions.

Two model families allow systematic investigation across model sizes within a fixed architecture. The Gemma~3 series \citep{gemma_gemma_2025} provides four instruction-tuned sizes: Gemma-3-1B-IT, Gemma-3-4B-IT, Gemma-3-12B-IT, and Gemma-3-27B-IT (1B--27B). The Qwen~3.5 series \citep{qwen_qwen35_2026} provides four thinking-capable sizes evaluated with thinking disabled: Qwen3.5-2B, Qwen3.5-4B, Qwen3.5-9B, and Qwen3.5-27B (2B--27B).

For cross-family comparison, we additionally evaluate Llama-3.1-8B-Instruct \citep{grattafiori_llama_2024} and Ministral-3-14B-Instruct \citep{liu_ministral_2026}. Together, these 10 models span four families and 1B--27B parameters.

\subsection{Task Prompts}
\label{app:prompts}

\paragraph{N-back task.}
Each trial is a multi-turn chat conversation with 50 turns. On each turn, the user presents one stimulus letter and the model returns one letter. Below we show the system prompt instantiated for 2-back. The inline 1-back and 2-back examples are random letter sequences regenerated per trial.

\begin{tcolorbox}[colback=gray!5!white, colframe=black!75!white, title=System Prompt (2-back), fonttitle=\bfseries]
\small
You are taking part in a working-memory experiment called an N-back task.

\smallskip
\textbf{How the task works:}\\
\quad -- You will see one letter at a time, written as ``X''.\\
\quad -- The first letter you see is turn $t = 0$, then $t = 1$, $t = 2$, and so on.\\
\quad -- On each turn, report the letter that appeared $N$ turns earlier.

\smallskip
\textbf{Response rule ($N = 2$):}\\
\quad -- If fewer than $N$ letters have appeared so far ($t < N$), respond with \texttt{-}.\\
\quad -- Otherwise ($t \geq N$), respond with the letter shown at turn $(t - N)$.\\
\quad -- Here are a 1-back and a 2-back example to help you understand:

\smallskip
\quad \textbf{1-back}\quad Stimuli: \texttt{BHTWYLMRGS}\quad Response: \texttt{-BHTWYLMRG}\\
\quad \textbf{2-back}\quad Stimuli: \texttt{FKDRMQNJPC}\quad Response: \texttt{-{}-FKDRMQNJ}

\smallskip
\textbf{Important:}\\
\quad -- To answer at turn $t$, use the letter presented at turn $t-N$; all other stimulus positions are irrelevant to the current response.\\
\quad -- Letters seen earlier than that are no longer relevant.\\
\quad -- Respond based only on the letter from $N$ turns ago.

\smallskip
\textbf{Response format (STRICT):}\\
\quad -- Respond with exactly one response per turn.\\
\quad -- The entire response must be a single token: A--Z or \texttt{-}.\\
\quad -- Do not include explanations, spaces, or extra text.

\smallskip
\textbf{Turn-by-turn example ($N = 2$):}\\
\quad User: A \quad Assistant: \texttt{-}\\
\quad User: B \quad Assistant: \texttt{-}\\
\quad User: C \quad Assistant: A\\
\quad User: D \quad Assistant: B\\
\quad User: E \quad Assistant: C
\end{tcolorbox}

\noindent During evaluation, each stimulus letter is delivered as a separate user message via the model's chat template, and the model produces one letter at that turn. Under teacher forcing, all earlier assistant responses are replaced by the ground-truth answers before the current turn is scored.

\begin{tcolorbox}[colback=gray!5!white, colframe=black!75!white, title=Multi-turn Evaluation (2-back), fonttitle=\bfseries]
\small
\textbf{<System>} [System prompt above]

\smallskip
\textbf{<User>} G\\
\textbf{<Assistant>} \texttt{-}

\smallskip
\textbf{<User>} K\\
\textbf{<Assistant>} \texttt{-}

\smallskip
\textbf{<User>} R\\
\textbf{<Assistant>} G

\smallskip
\textbf{<User>} F\\
\textbf{<Assistant>} K

\smallskip
\quad $\vdots$ \quad (50 turns total per trial)
\end{tcolorbox}

\paragraph{Letter-identity representations.}
For the representational analyses (\S\ref{sec:mechanistic}), we extract letter-identity representations (A--Z) under three conditions, recording hidden states across all transformer layers at each letter's token position.

\begin{tcolorbox}[colback=gray!5!white, colframe=black!75!white, title=Letter-Identity Representation Conditions, fonttitle=\bfseries]
\small

\textbf{1.\ Generic encoding} (\texttt{enc\_generic})\\
Each letter is presented as a user message under a generic system prompt.

\smallskip
\quad \textbf{<System>} You are a helpful assistant.\\
\quad \textbf{<User>} \{letter\}

\medskip
\textbf{2.\ N-back encoding} (\texttt{enc\_nback})\\
Same format, but under the N-back system prompt (with $N = 1$).

\smallskip
\quad \textbf{<System>} [N-back system prompt, $N = 1$]\\
\quad \textbf{<User>} \{letter\}

\medskip
\textbf{3.\ Generation averaged} (\texttt{gen\_averaged})\\
Each letter appears as an assistant-generated response. Five conversations with different user prompts are constructed and hidden states at the assistant token are averaged.

\smallskip
\quad \textbf{<System>} You are a helpful assistant.\\
\quad \textbf{<User>} \{prompt\}\\
\quad \textbf{<Assistant>} \{letter\}

\smallskip
Prompts: ``What's on your mind?'', ``What's the letter on your mind?'', ``What is the best letter you like?'', ``Give me a letter.'', ``Say a letter.''
\end{tcolorbox}

\subsection{Analysis Methods}
\label{app:analysis}

\subsubsection{Statistical summaries}
Unless noted otherwise, reported associations use Pearson correlation with two-sided $p$ values. Human reference curves are shown as inverse-probability-weighted, study-averaged estimates with 95\% bootstrap confidence intervals, and cross-model curves are summarized as mean $\pm$ SEM across models.

\subsubsection{Sample response-position hidden states}
For each trial and each evaluable turn $t \geq N$, we record hidden states from five evenly spaced transformer depths. We denote the response representation immediately before next-token prediction by $h_t^\ell$, where $t$ indexes the turn and $\ell$ the layer. For representation analyses, we use signed relative offsets $k \in \{-N,\ldots,0\}$ and define the letter-resolved mean $h_{k,c}^\ell = \mathbb{E}[h_t^\ell \mid x_{t+k}=c]$. Thus, $h_{0,c}^\ell$ represents the current item and $h_{-N,c}^\ell$ the target. All N-back expectations and averages below pool eligible instances across trials unless otherwise noted; we suppress the trial index for readability. The mechanistic figures use $h_t^\ell$, because this representation directly supports recall.

\subsubsection{Compute agreement-adjusted performance across memory load}
\begin{figure}[htbp]
\centering
\includegraphics[width=0.75\textwidth]{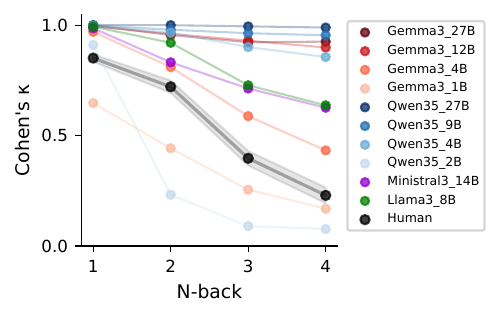}
\caption{\textbf{Teacher-forced performance across memory loads.} With ground-truth prior responses inserted into the context, performance remains higher than in autoregressive generation, but the decline with $N$ is preserved. Larger models shift the curve upward without changing its qualitative shape. The shaded region for humans indicates the 95\% bootstrap confidence interval of the aggregated human reference curve.}
\label{fig:capacity_tf}
\end{figure}

For the performance curves in Fig.~\ref{fig:perf_task} and Appendix Fig.~\ref{fig:capacity_tf}, each trial contributes only turns with a defined $N$-back target, $t \in \{N,\ldots,T-1\}$ with $T=50$. In the estimators below, $\mathcal{T}_N$ denotes all such turn instances pooled across trials. For each instance, the ground-truth label is the letter shown $N$ steps earlier, $y_t = x_{t-N}$, and the model prediction is the highest-probability letter,
\[
\hat{y}_t = \arg\max_{a \in \mathcal{A}} p_t(a),
\]
where $\mathcal{A}$ is the 26-letter response vocabulary. Let $|\mathcal{T}_N|$ denote the number of pooled evaluable turn instances. We then compute the observed agreement
\[
p_o = \frac{1}{|\mathcal{T}_N|}\sum_{t \in \mathcal{T}_N} \mathbf{1}[\hat{y}_t = y_t].
\]
We then adjust for non-uniform response marginals using Cohen's $\kappa$,
\[
\kappa = \frac{p_o - p_e}{1 - p_e}, \qquad
p_e = \sum_{a} p(y_t = a)\, p(\hat{y}_t = a),
\]
where $p_e$ is estimated from the empirical target and prediction frequencies within the same condition:
\[
p(y_t = a) = \frac{1}{|\mathcal{T}_N|}\sum_{t \in \mathcal{T}_N} \mathbf{1}[y_t = a], \qquad
p(\hat{y}_t = a) = \frac{1}{|\mathcal{T}_N|}\sum_{t \in \mathcal{T}_N} \mathbf{1}[\hat{y}_t = a].
\]
Operationally, for each model, generation mode, and $N$, we tabulate how often each letter appears as the target and as the model's argmax prediction across all evaluable turns, then substitute these empirical marginal frequencies into the formula above. This yields a chance-corrected measure of recall fidelity across $N$. For the human reference, we use the same correction with $p_e=0.5$.

\subsubsection{Compute the capacity--interference frontier and temporal retrieval kernels}
\begin{figure}[htbp]
\centering
\includegraphics[width=0.95\textwidth,clip]{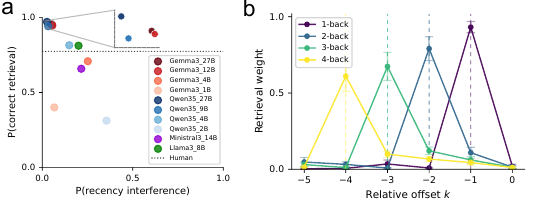}
\caption{\textbf{Teacher-forced retrieval patterns.} \textbf{(a)}~The capacity--recency relationship under teacher forcing; most models cluster near high target retrieval probability and low non-target recency mass, but the relationship is preserved. \textbf{(b)}~Teacher-forced retrieval kernels peak at the target offset but retain a recency tail that grows with memory load. Error bars indicate $\pm$1 SEM across models.}
\label{fig:retrieval_tf}
\end{figure}

For each offset $k \in \{-5,\ldots,0\}$, we quantify retrieval weight as the mean probability mass assigned to the memory item $x_{t+k}$,
\[
\rho_k = \frac{1}{|\mathcal{T}_{N,k}|}\sum_{t \in \mathcal{T}_{N,k}} p_t(x_{t+k}),
\]
where $\mathcal{T}_{N,k}$ contains evaluable turns for which $t+k$ is in range and, for $k \neq -N$, $x_{t+k} \neq x_{t-N}$. The temporal retrieval kernel in Fig.~\ref{fig:retrieval_ar} and Appendix Fig.~\ref{fig:retrieval_tf} plots $\rho_k$ as a function of $k$. The cross-model frontier uses $\rho_{-N}$ as target retrieval probability and $\sum_{k=-N+1}^{0} \rho_k$ as non-target recency mass, averaged across $N$.

\subsubsection{Analyze target-selective attention heads}
\label{app:attention_methods}
\begin{figure}[htbp]
\centering
\includegraphics[width=\textwidth]{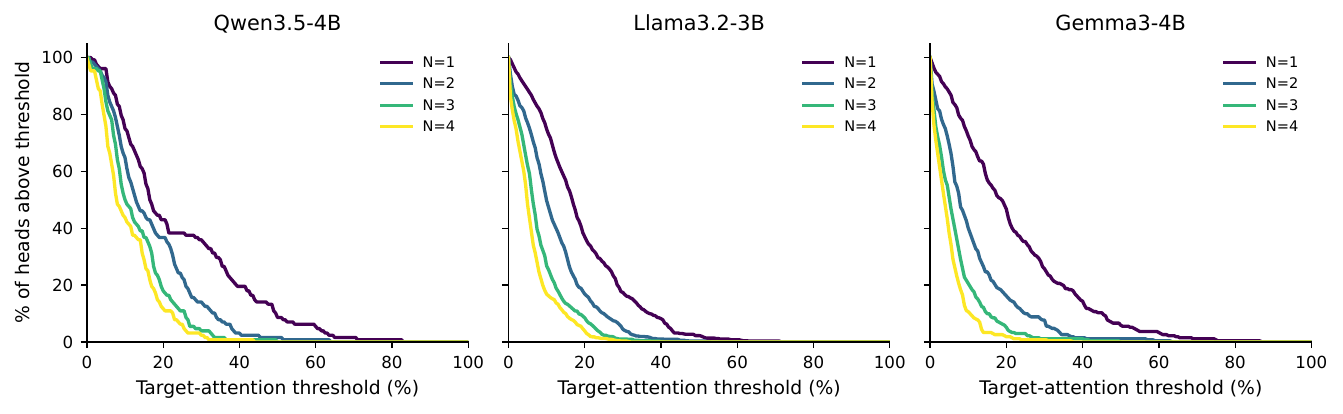}
\vspace{-5pt}
\caption{\textbf{Target-attention heads form a sparse tail.} For each model and $N$-back level, the curve reports the percentage of analyzed heads whose normalized attention to the target stimulus at offset $k=-N$ exceeds each threshold. Normalized attention is computed relative to observed-stimulus mass. Higher memory loads shift the survival curves downward, indicating that fewer heads allocate large fractions of stimulus-directed attention to the target.}
\label{fig:attention_head_survival}
\end{figure}

\begin{figure}[htbp]
\centering
\includegraphics[width=\textwidth]{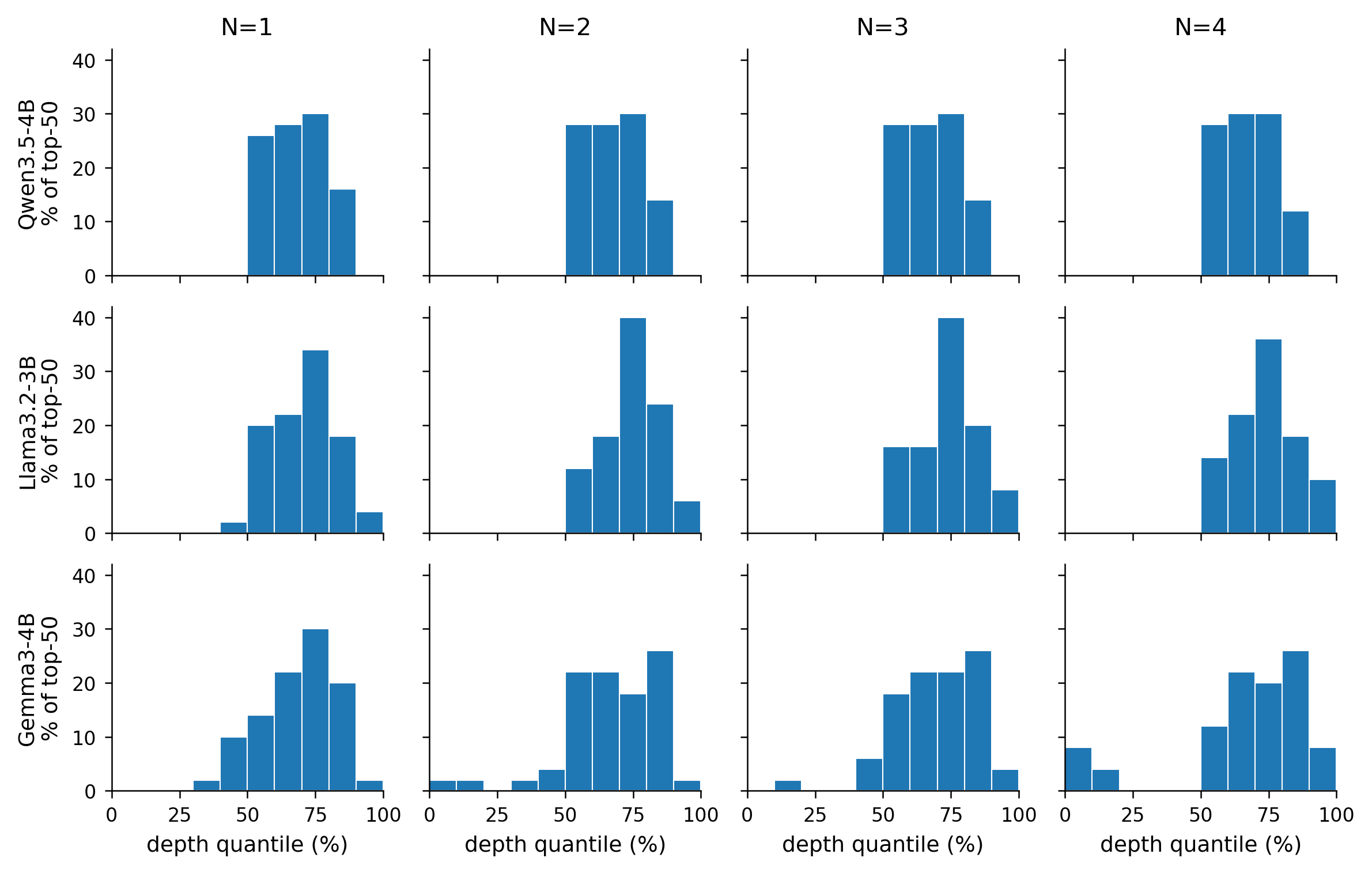}
\vspace{-5pt}
\caption{\textbf{Top target-attention heads concentrate in middle-to-late layers.} Histograms show the depth quantile of the top-50 heads ranked by normalized attention to the target at offset $k=-N$ for each model and $N$-back level. Across models, these heads are concentrated after the earliest layers, typically around the middle-to-late portion of the network.}
\label{fig:attention_head_depth}
\end{figure}

\begin{figure}[htbp]
\centering
\includegraphics[width=\textwidth]{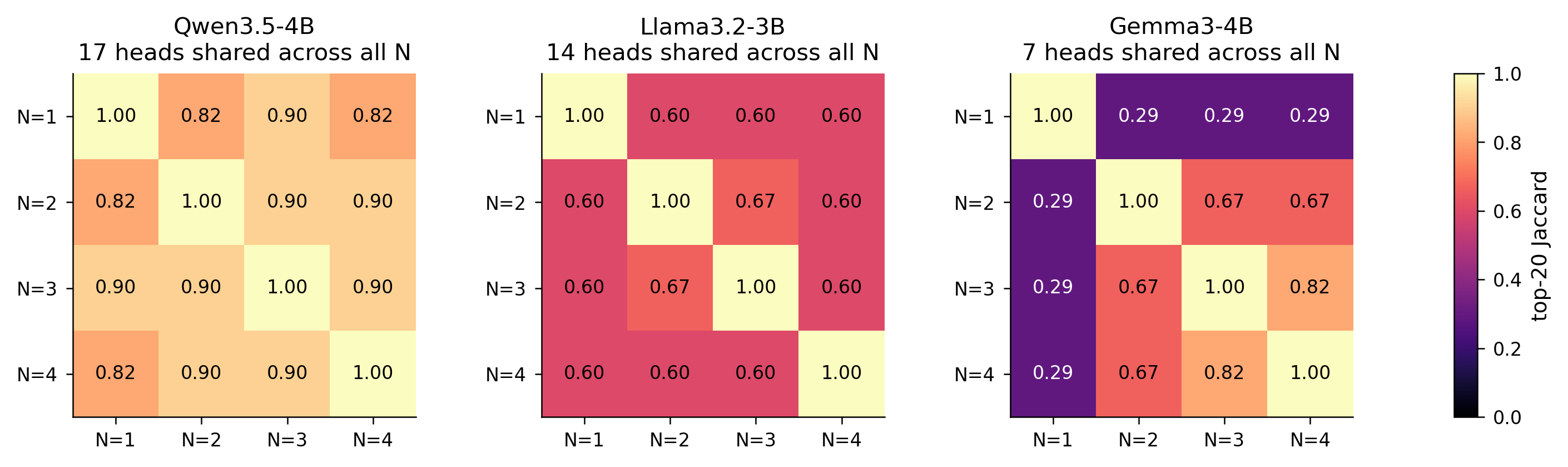}
\vspace{-5pt}
\caption{\textbf{Target-attention head sets are partially reused across memory loads.} Heatmaps show Jaccard overlap between the top-20 target-attention head sets for different $N$-back levels. The number in each title reports how many heads are shared across all four loads. Qwen3.5-4B and Llama3.2-3B show substantial reuse across $N$, whereas Gemma3-4B shows weaker overlap between 1-back and higher loads.}
\label{fig:attention_head_overlap}
\end{figure}

To measure whether self-attention routes a clean target pointer or a broader set of recent memory items, we replayed teacher-forced baseline trials and extracted explicit pairwise attention weights at the assistant-side query position immediately preceding each scored response. This analysis was restricted to models and layers that exposed standard attention tensors, yielding Qwen3.5-4B, Llama3.2-3B, and Gemma3-4B. For each evaluable turn $t \geq N$, layer $\ell$, and head $h$, let $A_{\ell,h,t}(i)$ denote attention from the response query to token position $i$, and let $e_j$ be the token position at which stimulus $x_j$ was encoded. We first computed the total attention mass assigned to observed stimulus tokens,
\[
M_{\ell,h,t} = \sum_{j=0}^{t} A_{\ell,h,t}(e_j),
\]
and then normalized the attention assigned to each signed stimulus offset $k$ by this stimulus-token mass,
\[
a_{\ell,h,t}(k) =
\frac{A_{\ell,h,t}(e_{t+k})}{\max(M_{\ell,h,t}, \epsilon)}.
\]
Here $\epsilon=10^{-12}$ only prevents division by zero, $k=0$ denotes the current stimulus, and $k=-N$ denotes the N-back target. For each model, $N$, layer, and head, we averaged $a_{\ell,h,t}(k)$ across evaluable turns and trials to obtain an offset profile $\bar{a}_{\ell,h}^{(N)}(k)$. Target attention was defined as $\bar{a}_{\ell,h}^{(N)}(-N)$, and target selectivity was this target attention minus the mean normalized attention over all other offsets $k \in \{-29,\ldots,0\}\setminus\{-N\}$. For visualization, the offset profiles were truncated to $k \in \{-8,\ldots,0\}$. Fig.~\ref{fig:attention_lag_profiles} shows the mean offset profiles of the top 10 heads ranked by target selectivity for each model and $N$. Appendix Fig.~\ref{fig:attention_head_survival} summarizes how many heads exceed each target-attention threshold, Appendix Fig.~\ref{fig:attention_head_depth} shows where the top-50 target-attention heads fall across layer depth, and Appendix Fig.~\ref{fig:attention_head_overlap} reports the Jaccard overlap among the top-20 target-attention head sets across $N$.

\subsubsection{Quantify lure, vocabulary, and transition effects on behavior}
For the behavioral manipulations summarized in Table~\ref{tab:interference_corr}, let $a(\cdot)$ denote mean accuracy under the indicated condition. We measure lure, vocabulary, and transition effects as
\[
\Delta_{\mathrm{lure}} = \frac{1}{2}\big(a_{\mathrm{left}} + a_{\mathrm{right}}\big) - a_{\mathrm{base}}, \qquad
\tilde{a}_m = \frac{a_m - 1/m}{1 - 1/m},
\]
\[
\Delta_{\mathrm{vocab}} = \tilde{a}_{10} - \tilde{a}_{26}, \qquad
\Delta_{\mathrm{tran}} = a_{10,\,p_{\mathrm{tran}}=0.8} - a_{10,\,p_{\mathrm{tran}}=0}.
\]
Here, $a_{\mathrm{left}}$ and $a_{\mathrm{right}}$ are the mean accuracies for the $N\!-\!1$ and $N\!+\!1$ lure conditions, $a_{\mathrm{base}}$ is the baseline accuracy, and $\tilde{a}_m$ is the chance-corrected accuracy for an $m$-letter response set. These contrasts quantify, respectively, sensitivity to positional lures, reduced vocabulary size, and predictable transition structure.

\subsubsection{Construct letter-identity representations}
To measure how strongly working memory states preserve letter identity, we construct letter-identity representations for each letter under three contexts: generic stimulus encoding, stimulus encoding under the N-back instruction prompt, and isolated letter generation as an assistant response. For each context and layer, the letter-identity representation for letter $c$ is the mean hidden-state vector $s_c^\ell$ across repeated presentations. Together, these contexts provide reference estimates of letter-identity geometry across encoding and generation settings.

\subsubsection{Measure residual letter-identity information by alignment to letter-identity representations}
To evaluate how much the response representation still preserves the identity of the current stimulus, we use the current-item representation $h_{0,c}^\ell$. Alignment with the corresponding letter-identity representation is then
\[
\mathrm{Align}_\ell = \frac{1}{26}\sum_{c} 
\frac{h_{0,c}^\ell}{\|h_{0,c}^\ell\|} \cdot
\frac{s_c^\ell}{\|s_c^\ell\|}.
\]
This quantity is high when the current-item representation for letter $c$ remains close to that letter's task-independent letter-identity representation and low when that identity has been suppressed or transformed. The analysis underlies Fig.~\ref{fig:neural_control}a and the letter-alignment correlations in Table~\ref{tab:trial_corr}. For summary plots, we average alignment across the three letter-identity representations described above.

\subsubsection{Measure residual letter-identity information by decoding from response representations}
As a complementary test of explicit current-letter information, we quantify how well the set of current-item representations $\{h_{0,c}^\ell\}_c$ remains separable across letters. Operationally, we decode $x_t$ directly from each response representation by nearest-centroid classification. We fit the centroids on training trials and evaluate held-out trials using trial-wise cross-validation. Suppressing the fold index, each held-out turn is assigned
\[
\hat{c}_{t,\ell} = \arg\max_{c} \cos\!\big(h_t^\ell,\, h_{0,c}^\ell\big).
\]
Decoding accuracy is averaged across held-out turns to produce the decoding panel of Fig.~\ref{fig:neural_control}. Higher accuracy indicates that current-letter identity remains explicitly recoverable from the response representation $h_t^\ell$ and that the current-item representations $\{h_{0,c}^\ell\}_c$ remain well separated; lower accuracy indicates that this content has been stripped away. These values also enter the decodability correlations in Table~\ref{tab:trial_corr}.

\subsubsection{Measure relative-position disentangling by similarity among memory-item representations}
To assess whether the network reorganizes memory by temporal position, we subtract the global response mean $\bar{h}^\ell = \mathbb{E}_t[h_t^\ell]$ from the letter-resolved relative-position means, yielding $\tilde{h}_{k,c}^\ell = h_{k,c}^\ell - \bar{h}^\ell$. Pairwise positional similarity is
\[
S_{k,k',\ell} = \frac{1}{26}\sum_c \cos(\tilde{h}_{k,c}^\ell, \tilde{h}_{k',c}^\ell),
\]
and the summary disentangling measure is the mean absolute off-diagonal similarity $\mathbb{E}_{k \neq k'} |S_{k,k',\ell}|$. Lower values indicate that different relative-position representations form more distinct representational directions. This quantity underlies Fig.~\ref{fig:neural_control}c and the subspace-similarity correlations in Table~\ref{tab:trial_corr}.

\subsubsection{Measure target readiness by alignment to the correct readout direction}
To investigate when the target representation becomes directly readable by the output layer, we define for each letter $c$ a letter-specific readout direction $r_c$ by averaging the model's output-weight vectors associated with that letter. Readout alignment at relative offset $k$ is then
\[
R_{k,\ell} = \frac{1}{26}\sum_c \cos(\tilde{h}_{k,c}^\ell, r_c).
\]
The final panel of Fig.~\ref{fig:neural_control} reports $R_{-N,\ell}$, the mean cosine alignment between centered target-position states and their corresponding letter-specific readout directions. Larger values indicate that the target representation is already oriented for output. These values also enter the target-alignment correlations in Table~\ref{tab:trial_corr}.

\subsubsection{Relate representational metrics to behavioral accuracy}
For Table~\ref{tab:trial_corr}, we relate each representational statistic --- letter alignment, decoding accuracy, mean off-diagonal positional similarity, or target-readout alignment --- to recall success within each model. Position-wise alignment and geometry use position-specific means, whereas position-wise decodability is held-out accuracy at each position using centroids fitted on training trials. The sequence-position and fold indices are suppressed above for readability. For a given model, we then correlate each statistic across sequence positions, pooling $N = 1$--$4$, with the mean target log-probability at each position, a graded index of recall. These per-model Pearson coefficients are aggregated across the ten models by Fisher $z$-transformation, with significance combined using Fisher's method; autoregressive and teacher-forced evaluation are analyzed separately.

\subsubsection{Causally remove letter-identity directions from the residual stream}
\label{app:causal_remove}
To test whether residual letter-identity information contributes causally to N-back errors, we intervened directly on the residual stream during evaluation. We first averaged the letter-identity representations from the three contexts described above and, for each layer, performed an SVD on the centered 26-letter centroid matrix. This yielded layer-specific principal directions spanning the dominant letter-identity subspace. During a forward pass, we inserted a hook at one of the two earliest sampled transformer depths and moved the hidden state toward the mean projection along the selected direction, replacing letter-specific variation while preserving the overall presence of signal along that direction. The hook acts only at the answer-generation positions, namely the assistant-side hidden states immediately preceding each response token, and N-back accuracy is evaluated from the logits at those same positions. Formally, for hidden state $h$, selected basis direction $B$, mean projection $\mu_{\mathrm{proj}}$, and intervention strength $\alpha$, we applied $h \leftarrow h - \alpha(\mathrm{proj}_B(h) - \mu_{\mathrm{proj}})$, with $\alpha \in \{0.3, 0.5, 1.0\}$. Here, $\mathrm{proj}_B(h)$ denotes the orthogonal projection of $h$ onto $B$, and $\mu_{\mathrm{proj}}$ is the mean projection of the 26 context-averaged letter centroids at the intervened layer.

We evaluated this intervention with seed-controlled trials so that each intervention was compared against the same sampled sequences as its baseline. For each model and each $N \in \{1,2,3,4\}$, we swept individual principal directions, intervention strengths, and the two earliest sampled depths, using 50 trials per condition. Due to high computational cost under autoregressive generation, we performed this analysis on a computationally tractable subset of models spanning a broad performance range rather than on the full model set. We then summarized this sweep across models.

Fig.~\ref{fig:causal_removal} reports the model-level summary used in the main text. For each model and $N$-back level, we compared baseline accuracy with the maximum post-intervention accuracy in the sweep and reported the larger value, thereby clamping negative gains to zero. The plotted bars average these baseline and best-intervention accuracies over $N = 1$--$4$. This best-in-sweep summary measures whether any tested suppression of letter-identity information improves retrieval within the evaluated model and load conditions. Estimating transfer requires selecting the intervention on separate trials and applying a fixed configuration across held-out models or memory loads.

\subsection{Extended Figures for Attention Routing}
\label{app:attention_figures}

\begin{table}[htb]
\centering
\caption{\textbf{N-back accuracy by model and generation mode.} Accuracy (proportion correct) on the baseline condition ($|\mathcal{V}|=26$, uniform) for $N = 1$--$4$ under autoregressive generation and teacher forcing. Models are grouped by family and sorted by parameter count.}
\label{tab:nback_accuracy}
\small
\begin{tabular}{l cccc cccc}
\toprule
& \multicolumn{4}{c}{\textbf{Autoregressive}} & \multicolumn{4}{c}{\textbf{Teacher Forcing}} \\
\cmidrule(lr){2-5} \cmidrule(lr){6-9}
\textbf{Model} & 1 & 2 & 3 & 4 & 1 & 2 & 3 & 4 \\
\midrule
Qwen3.5 2B   & 0.11 & 0.05 & 0.05 & 0.05 & 0.91 & 0.26 & 0.12 & 0.11 \\
Qwen3.5 4B   & 0.92 & 0.50 & 0.26 & 0.21 & 1.00 & 0.97 & 0.90 & 0.86 \\
Qwen3.5 9B   & 0.99 & 0.44 & 0.45 & 0.41 & 1.00 & 0.98 & 0.97 & 0.96 \\
Qwen3.5 27B  & 1.00 & 0.99 & 0.88 & 0.84 & 1.00 & 1.00 & 0.99 & 0.99 \\
\midrule
Gemma3 1B    & 0.05 & 0.04 & 0.05 & 0.03 & 0.66 & 0.46 & 0.28 & 0.20 \\
Gemma3 4B    & 0.71 & 0.03 & 0.04 & 0.05 & 0.97 & 0.82 & 0.60 & 0.45 \\
Gemma3 12B   & 0.95 & 0.36 & 0.10 & 0.05 & 1.00 & 0.96 & 0.93 & 0.90 \\
Gemma3 27B   & 1.00 & 0.30 & 0.13 & 0.08 & 1.00 & 0.96 & 0.93 & 0.93 \\
\midrule
Llama-3.1-8B-Instruct & 0.98 & 0.30 & 0.10 & 0.08 & 0.99 & 0.92 & 0.74 & 0.65 \\
Ministral3 14B & 0.95 & 0.13 & 0.15 & 0.12 & 0.99 & 0.84 & 0.72 & 0.64 \\
\bottomrule
\end{tabular}
\end{table}

\subsection{Extended Figures for Teacher Forcing}
\label{app:extended_tf}

This subsection collects the teacher-forced counterparts to the main-text autoregressive figures. Teacher forcing removes error propagation from previous responses, allowing us to test whether the same qualitative patterns persist when earlier answers are fixed to ground truth.

\begin{figure}[htbp]
\centering
\includegraphics[width=0.95\textwidth]{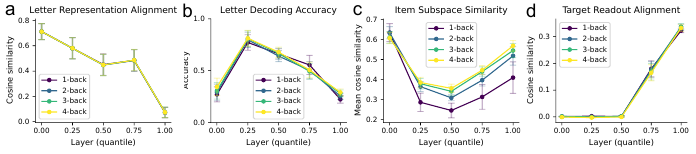}
\caption{\textbf{Teacher-forced neural control signatures.} \textbf{(a)}~Alignment with letter-identity representations decreases across network layers. \textbf{(b)}~Explicit letter decoding peaks in early layers before declining. \textbf{(c)}~Relative-position representations disentangle in middle layers but re-overlap near the output stage, exhibiting greater overlap under higher memory loads ($N$). \textbf{(d)}~Alignment between target representations and output readout directions emerges strictly in late layers. The sequence matches the autoregressive result while removing contamination from self-generated errors. Shaded bands indicate $\pm$1 SEM across models.}
\label{fig:neural_control_tf}
\end{figure}

\end{document}